\documentclass[]{PHMSociety}

\usepackage{graphicx}
\usepackage{amsmath}
\usepackage{amsfonts}
\usepackage{amssymb}
\usepackage{subcaption}
\usepackage{float}
\usepackage[hyphens]{url}
\usepackage[linesnumbered,ruled,vlined]{algorithm2e}

\begin{document}

\title{Complementary Meta-Reinforcement Learning for Fault-Adaptive Control}

\author{%
	Ibrahim Ahmed\authorNumber{1}, Marcos Quiñones-Grueiro\authorNumber{2}, and Gautam Biswas\authorNumber{3}
}

\address{
	\affiliation{{1, 2, 3}}{Vanderbilt University, Nashville TN, 37209 USA}{ 
		{\email{ibrahim.ahmed@vanderbilt.edu}}\\
		{\email{gautam.biswas@vanderbilt.edu}}\\ 
		{\email{marcos.quinones@vanderbilt.edu}}
		} 
}

\maketitle

\phmLicenseFootnote{Ibrahim Ahmed}

\begin{abstract}
Faults are endemic to all systems. Adaptive fault-tolerant control maintains degraded performance when faults occur as opposed to unsafe conditions or catastrophic events. In systems with abrupt faults and strict time constraints, it is imperative for control to adapt quickly to system changes to maintain system operations. We present a meta-reinforcement learning approach that quickly adapts its control policy to changing conditions. The approach builds upon model-agnostic meta learning (MAML). The controller maintains a complement of prior policies learned under system faults. This ``library" is evaluated on a system after a new fault to initialize the new policy. This contrasts with MAML, where the controller derives intermediate policies anew, sampled from a distribution of similar systems, to initialize a new policy. Our approach improves sample efficiency of the reinforcement learning process. We evaluate our approach on an aircraft fuel transfer system under abrupt faults.
\end{abstract}

\section{Introduction} \label{sec:intro}

No physical system is immune to degradation, changing environments, and faults. Since such situations can occur during operation, it is important the system respond to these changes in a way that the system continues to operate, be it in a degraded manner. This ensures safety and cost-effectiveness through less down-time. Fault-tolerant control (FTC) \cite{ftcbook} seeks to keep a faulty system operating, but within an acceptable margin of sub-optimal performance. This relaxes the constraints on the designers to make a system completely fail-safe and allows for a tradeoff between design and operating costs.

Data-driven approaches to FTC \cite{datadrivenftc1, datadrivenftc2} exploit the preponderance of data collected from system operations. They generate models that avoid the need for time-consuming and accurate physics-based simulations of system dynamics to analyze and respond to different situations that may occur in the system. However, such methods depend on the data to span the breadth of operating conditions, and the model has to contain sufficient detail to capture multiple operating modes. This represents another compromise between design and operating costs.

In many cases, systems are complex, the number of possible faults are large, and faults that have not been seen before can occur during operations. There may not be precedent in the data to model such behaviors. A data-driven control approach will not then have ``ground truth" to learn from and recall a sufficiently optimal control policy. Reinforcement learning (RL) is a semi-supervised approach to machine learning. It forfeits dependence on labelled ground truth and instead relies on accumulated feedback (i.e. experience gained) from a sequence of actions to determine a globally optimal policy. This ability to  learn during operations alleviates design time effort and costs.

RL relies on gathered experience to accurately evaluate actions. This can be represented as a dynamic programming problem \cite{dynamicprogramming} that typically has a closed-form solution, but for large systems, suffers from the curse of dimensionality. Advancements to RL have used function approximations of values to overcome the computational intractability of the problem \cite{valueapproximation1,valueapproximation2}. However, the dependence on data to learn such approximations limits how fast and how accurately a RL-based controller can accommodate faults.

In our past work \cite{previouswork}, we developed data-driven models to supplement experience with the real environment and simulate faults. In this work, we employ meta-RL for faster adaption of the RL algorithm to collected data samples. Our approach is not dependent on the time-consuming step of a data-driven model being learned first, however one can be used. The next section provides a background on RL and meta-RL. Section \ref{sec:cmaml} describes our approach, and section \ref{sec:exp} evaluates it on a simulation of a fuel-transfer system. Finally, section \ref{sec:related} places our work in the context of extant research.
\section{Preliminaries} \label{sec:preliminaries}

This section briefly introduces the RL approach and then discusses Model Agnostic Meta Learning in the context of RL-based control.

\subsection{Reinforcement Learning}

Reinforcement Learning (RL) is a semi-supervised approach to machine learning. A RL problem consists of a controller interacting with its environment. The environment can be modeled as a single Markov Decision Process (MDP) sampled from a population of available processes, $p \sim P$. At a time $t$ the controller perceives the environment's state $x_t \in X$, and uses its policy $\pi: X \rightarrow U$ to take an action $u_t \in U$. The environment goes into a new state $x_{t+1}$ governed by its transition function $T: X \times U \rightarrow X$ and emits a reward signal $r_t \in \mathbb{R}$, defined by the function $R: X \times U \times X \rightarrow \mathbb{R}$. The combination of $(X, U, T, R)$ constitutes a MDP, $p\in P$.

The goal of a RL is to maximize the return, $J_\pi(x, u)$,  which represents the total discounted cumulative reward for an action from each state when a policy, $\pi$ is followed. A discount factor $\gamma \in [0, 1]$ is used to weigh immediate rewards over delayed rewards and to ensure convergence of the discounted reward  function. The maximum future discounted reward for an action from a state is its value $V: X \times U \rightarrow \mathbb{R}$:

\begin{align}
    V(x_T, u_T) &= \max_{\pi} J_\pi(x_T, u_T) \nonumber \\
                &= \max_{\pi} \Sigma_{t=T}^\infty \gamma^{t-T} \cdot r_t  \nonumber\\
                &= r_T + \gamma \cdot \max_{u_T+1} V(x_{T+1}, u_{T+1})
\end{align}

Policy gradient algorithms \cite{policygradient} in RL parametrize $\pi$ with parameters $\theta$, i.e. $\pi_\theta$. The parameters $\theta$ are be the weights of a model representing the policy, for e.g. neural network. During training, they directly learn $\pi_\theta$ by implicitly optimizing for $V$ using gradient ascent on the gain function $G \leftarrow \mathbb{E}[ J_\pi(x, u)]$. Gradient ascent produces iterative updates to $\theta$ the size of which is determined by the learning rate $\alpha \in [0, 1]$.

Parameter updates at each iteration are dependent on experienced rewards under the latest policy. This is known as \textit{on-policy} RL. This approach is sample inefficient because new trajectories of interactions need to be obtained for each version of $\theta$. A way around this is to use \textit{importance sampling} in the gain function. By modeling the policy as a stochastic function over actions, $\pi_\theta (x \mid u)$, the relative probabilities, known as importance ratios, of the same trajectory under different policies can be obtained. Thus, the gain function can reuse the same batch of experiences to update the current iteration of parameters $\theta'$ by weighing cumulative rewards. Equation \ref{eqn:importance} shows how importance sampling reuses experiences collected under $\theta_k$ for the next iterations of policy parameters $\theta_{k+i}: i \geq 0$. The learning rate is $\alpha$.

\begin{align}
    G &= \mathbb{E}_{ x_0 \sim X}
    \left(\Pi_{t=0}^\infty \frac{\pi_{\theta_{k+i}} (x_t | u_t)}{\pi_{\theta_k} (x_t | u_t)}\right) J_{\pi_{\theta_k}}(x_0, u_0) \nonumber \\
    \theta_{k+i+1} &= \theta_k + \alpha \cdot \nabla_{\theta_{k+i}} G
    \label{eqn:importance}
\end{align}

Large gradient updates may cause the next iteration of $\pi_\theta$ to overshoot, thus missing the optimum, causing the learning process to diverge altogether. Proximal Policy Optimization (PPO) \cite{ppo} clips the size of gradient updates by restricting the importance ratios between iterations. Thus a policy does not drastically change between updates. We use PPO in this work to learn the control policy under fault conditions.

\subsection{Model-Agnostic Meta Learning}

Meta-learning seeks to speed up a machine learning process through introspection. Essentially, it learns how to learn. In a RL context, meta-learning seeks to quickly adapt a policy trained on one process to another.

Model-agnostic Meta Learning (MAML) \cite{maml} speeds up the optimization of any model learned through gradient updates. It does so by running an inner \textit{introspective} loop within each iteration of a gradient update to the model's parameters, which is designated as the outer loop. In the inner loop, variants of the process are sampled as $p^i \sim P$. The current model parameters $\theta$ are then optimized by training for several interactions on each $p^i$ using gradient ascent to yield $\theta^i$. At the end of the inner loop, gradients on a test set of interactions are computed. In the outer loop, the update to $\theta$ is a weighed aggregate of the test gradients from the inner loop. That is, the training step for the outer loop is based on the test step of the inner loop.

\begin{algorithm}[h]
\KwIn{parameters $\theta_k$, MDPs $P$, learning rates $\alpha_{in}, \alpha_{out}$, iterations $K_{in}, K_{out}$}
\Begin{
Set $\theta' \leftarrow \theta_k$\;
\For{$k_{out} = 1$ \KwTo $K_{out}$}{
    Sample MDPs $p^i \sim P$\;
    \For{all $p^i$}{
        Set $\theta^i \leftarrow \theta'$\;
        \For{$k_{in} = 1$ \KwTo $K_{in}$}{
            Sample training trajectories $\mathcal{M}^i$ from $p^i$\;
            Calculate gain function from $\mathcal{M}^i$\;
            Update $\theta^i \leftarrow \theta^i + \alpha_{in} \cdot \nabla_{\theta^i} G$\;
        }
        Sample test trajectories from $p^i$\;
        Calculate test gain $G^i$ on sample;
    }
    Update $\theta' \leftarrow \theta' + \alpha_{out} \cdot \Sigma_i \nabla_{\theta'} G^i$\;
}
\Return{$\theta'$}
}

\caption{Model-agnostic meta-learning\label{alg:maml}}
\end{algorithm}
\section{Complementary Meta-Reinforcement \\Learning} \label{sec:cmaml}

\subsection{Problem Formulation}

\begin{figure}[]
\begin{center}
\includegraphics[width=8.2cm]{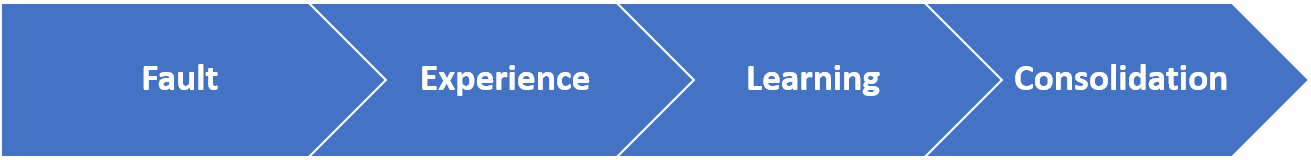}
\caption{}
\label{fig:framework}
\end{center}
\end{figure}

The problem of the controller is thus: to exploit its past experiences with different processes, and sparse interactions under new process dynamics $p'$ to quickly converge to a locally optimal policy. The proposed approach for adaptive control operates under the framework depicted in figure \ref{fig:framework}. The adaption pipeline can either be preempted by fault detection, or happen periodically.

The adaption step begins with a \textit{fault}. The fault is abrupt, causing a discontinuous change in process dynamics $p \rightarrow p'$. The MDP representing the system has changed. In the aftermath of a fault, a controller continues to interact with $p'$ and records states, actions, and rewards in a memory buffer $\mathcal{M}$ using its current policy parameters $\theta_k$. Once sufficient interactions $t_{update}$ have been buffered, the controller attempts to initialize new parameters $\theta'$ from its memory, and then fine-tunes them to $\theta_{k+1}$ by interacting with the new process. Once learning is complete, the controller consolidates the newly learned policy with its prior policies. Thus, when a new fault occurs, it is able to exploit its past experience and adapt faster.

The \textit{learning} phase consists of two stages: the meta-update using the memory, followed by iterations of any choice of a gradient-based reinforcement learning algorithm on the new process. During the meta-update, the controller uses its consolidated prior experience to initialize new policy parameters. The controller can also generate a data-driven model of the system to supplement sample inefficiency of RL. After that, the parameters are iteratively updated by the RL algorithm through interactions with the actual system.

\textit{Consolidation} of knowledge happens via maintaining a complement of prior policies $\mathcal{C} = \{\theta \mid \pi_\theta\}$. The set of policies is periodically pruned to ensure that they capture diverse behavior but are small enough to evaluate within time constraints.

\subsection{Policy meta-update}

\begin{figure}[]
\begin{center}
\includegraphics[width=5cm]{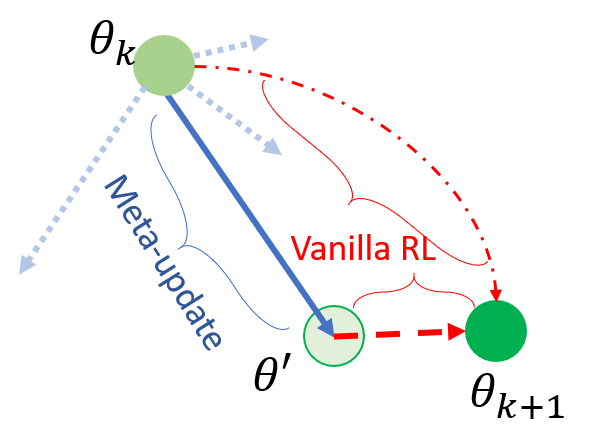}    
\caption{The meta-update initializes policy parameters closer to an optimum, after which RL converges faster to a solution. The meta-update depends on the aggregate gradients of policies in the complement. The gradients are calculated from samples from a data-driven process model updated from a buffer of recent experiences, or the buffer itself. The meta-update step from $\theta_k$ to $\theta'$ is described in algorithm \ref{alg:metarl}.}
\label{fig:metaupdate}
\end{center}
\end{figure}

Our approach mirrors MAML in that there is an outer update loop for the main policy parameters. It depends on the gradients of the test error on the inner loop. We diverge in our formulation of the inner loop. In MAML the inner loop samples random processes from a population $p^i \sim P$ defining the MDP. It uses those samples to derive intermediate parameters $\theta^i$ from the single starting parameter $\theta_k$. We forego sampling processes anew to derive such intermediate parameters, and instead exploit the history of the controller's experience. In other words, MAML evaluates multiple processes on a single set of parameters. We propose to evaluate a single process on multiple sets of parameters.

Prior to the meta-update, a memory $\mathcal{M}$ of interactions under the new process is buffered. The meta-update step assumes a complement $\mathcal{C} = \{\theta \mid \pi_\theta\}$ of prior policies trained on the system under different faults. This foregoes the need of sampling an altogether new set of processes for the meta-update. The complement of polices is then trained for a few steps $K_{in}$ to yield an updated set of meta-parameters. Finally, the test error of the meta-parameters on the process is used to update the outer loop's policy parameters.

Optionally, as a guard against a sub-optimal initialization $\theta_k \rightarrow \theta'$, $\theta_k$ is also concurrently updated using standard RL without meta learning to a baseline parameter $\theta^b_{k_{out}}$ for each iteration $k_{out}$ of the outer update loop. Finally, the meta-learned parameters and baseline parameters are evaluated on a provided process model $p_m$. Whichever performs better is returned as the new initialization $\theta'$.

\begin{figure}[]
\begin{center}
\includegraphics[width=7cm]{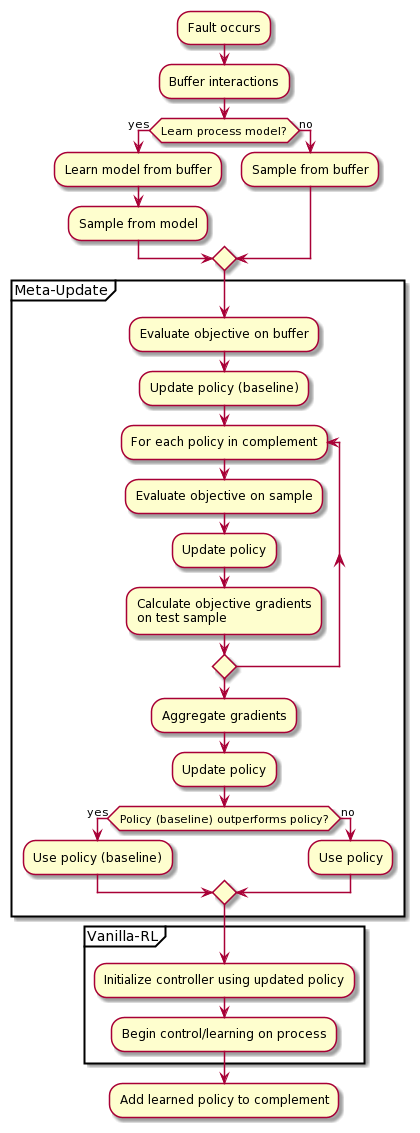}    
\caption{An overview of the complementary MAML algorithm in a FTC context. The meta-update step initializes the policy based on a complement of policies evaluated on the new process.}
\label{fig:algorithm}
\end{center}
\end{figure}

Evaluating policies from $\theta^i \in \mathcal{C}$ necessitates new interactions with the changed process $p'$. This can be achieved by learning a data-driven model $p_m$ of the process using $\mathcal{M}$. However, this introduces an additional computational load  on the meta-update step. An alternative approach, already inherent in PPO, is to forego a model altogether and instead use importance sampling (equation \ref{eqn:importance}) to adjust the gain with respect to $\theta^i$. With importance sampling, the returns already calculated on $p'$ under $\theta_k$ stored in $\mathcal{M}$ can be weighed by the relative probabilities of actions under $\theta^i$. This process is delineated in algorithm \ref{alg:metarl} and figure \ref{fig:algorithm}.

\begin{algorithm}[]
\SetKwInOut{KwOptional}{Optional}

\KwIn{parameters $\theta_k$, memory $\mathcal{M}$, learning rates $\alpha_{in}, \alpha_{out}$, iterations $K_{in}, K_{out}$}
\KwOptional{policy complement $\mathcal{C} = \{\varnothing\}$, process model $p_m=\varnothing$}

\Begin{
\If{$p_m \neq \varnothing$ }{
    Update $p_m$ from $\mathcal{M}$\;
    Sample trajectories from $p_m$, using policy\;
} \Else{
    Sample trajectories from $\mathcal{M}$, discarding policy\;
}
Set meta-updated params $\theta' \leftarrow \theta_k$\;
Set baseline params $\theta^b \leftarrow \theta_k$\;

\For{$k_{out} = 1$ \KwTo $K_{out}$}{
    Calculate gain from $\mathcal{M}$\;
    Update $\theta^b$ using $\alpha_{out}$\;
    
    \For{all $\theta^i$ in $\mathcal{C}$}{
        Sample trajectories and calculate gain\;
        Update $\theta^i$\ using $\alpha_{in}$\;
        Calculate test gain $G^i$\;
    }
    Update $\theta' \leftarrow \theta' + \alpha_{out} \cdot \Sigma_i \nabla_{\theta'} G^i$\;
}
Calculate $J_{\pi_{\theta'}}, J_{\pi_{\theta^b}}$ from $p_m$\;
\If{$J_{\pi_{\theta'}} < J_{\pi_{\theta^b}}$}{
\Return{$\theta' \gets \theta^b$}
} \Else{
    \Return{$\theta'$}
}
}
\caption{Complementary meta-RL\label{alg:metarl}}
\end{algorithm}

\subsection{Population of complement}

The final step of the approach is to store the newly learned parameters for future reference. The complement of policies should be populated with policies such that it maximally spans the parameter space. Policies should be different enough so that the meta-update has a greater likelihood of adapting to novel faults. The difference between policies is evaluated on the memory of interactions collected by the controller. Each policy in $\mathcal{C}$ generates a probability for actions stored in $\mathcal{M}$. KL-divergence between the probabilities is used as a metric of difference. The total divergence of each policy from the rest of the complement becomes a score of a policy's uniqueness. Given a complement size $|\mathcal{C}| \gets s$, the $s$ most unique policies are kept as new members of $\mathcal{C}$. Algorithm \ref{alg:complement} goes through the process of selecting between the existing and newly learned policies to update $\mathcal{C}$.

\begin{algorithm}[]
\KwIn{policy complement $\mathcal{C}$, complement size $s$, memory $\mathcal{M}$}
\Begin{
Initialize divergence matrix $D=[0]^{|C|\times|C|}$\;
\For{$\theta_1, \theta_2 \text{ in Permute}(\mathcal{C})$}{
    Action probabilities $p_1, p_2 = \pi_{\theta_1}(\mathcal{M}), \pi_{\theta_2}(\mathcal{M})$\;
    KL-Divergence $d = \Sigma p_1 \cdot \log (p2 / p1)$\;
    $D[\theta_1, \theta_2] = d$\;
    Sum each row of $D$ for total divergence $D_T^{|C|\times 1}$\;
    Most divergent parameters $\mathcal{C} \leftarrow \text{Sort}(\mathcal{C})\text{ by }D_T$\;
    \Return{First $s$ parameters from $\mathcal{C}$}
}
}

\caption{Populating complement of policies\label{alg:complement}}
\end{algorithm}
\section{Experiments} \label{sec:exp}

The algorithm was evaluated on a simulation of a fuel transfer system of an aircraft. The system is defined in greater detail in \cite{previouswork}. The objective is to maintain center of gravity, variance in fuel distribution, and closed valves to avoid unnecessary mass transfer. Faults can include increased valve resistances leading to low flow rates, and increased fuel consumption due engine faults.

A controller was first trained for 50,000 steps on the nominal system. At the beginning of a trial, a random fault occurred and the controller accumulated experience in memory $\mathcal{M}$. The controller then employed the meta-update step in algorithm \ref{alg:metarl} to initialize new policy parameters. Following that, the RL algorithm continued to learn on the new system. As a baseline, an RL controller was trained for $|\mathcal{C}| \times K_{in} \times K_{out}$ iterations on $p_m$, when $p_m$ was provided, followed by learning on the new system $p'$. For all experiments, a first-order approximation of gradients $\nabla_{\theta'}G^i$ as documented in \cite{maml} is used.

\begin{figure}[ht]
\begin{center}
\includegraphics[width=5cm]{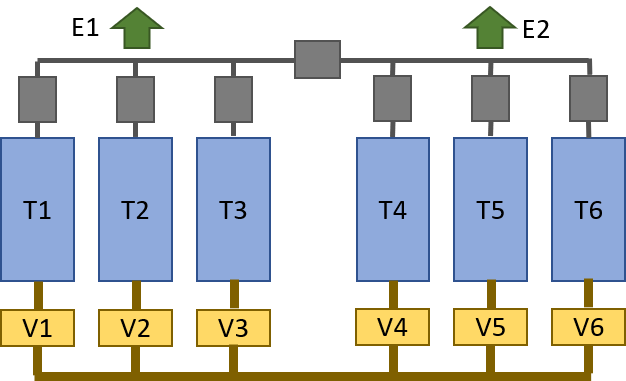}    
\caption{Simplified fuel system schematics. The controller manages valves and can observe fuel tank levels. Net outflow to engines via pumps is controlled independently. Pumps drain tanks innermost first.}
\label{fig:fuelsystem}
\end{center}
\end{figure}

First, the controller was tested with an empty complement of policies. Second, a complement of 3 policies under simulated faults on the system was generated. The complement was trained on faults in tanks 1, 3, and 5 and no engine faults. In both cases, the controller was tested on the system under random novel faults. The controller was allowed to adapt solely from buffered experiences after a fault, without learning a new environment model.

\begin{figure}[h]
\begin{center}
\includegraphics[width=8.2cm]{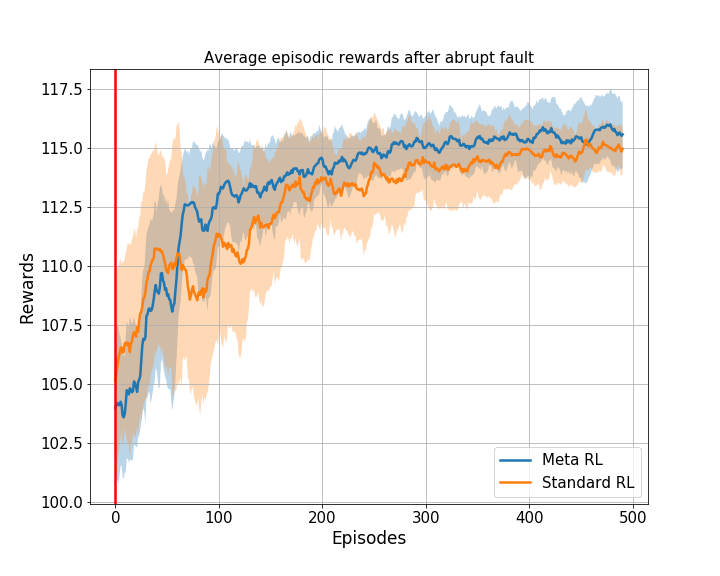}
\caption{Episodic rewards after abrupt fault when there is no complement of policies available to the controller.}
\label{fig:nocomplement}
\end{center}
\end{figure}

\begin{figure}[h]
\begin{center}
\includegraphics[width=8.2cm]{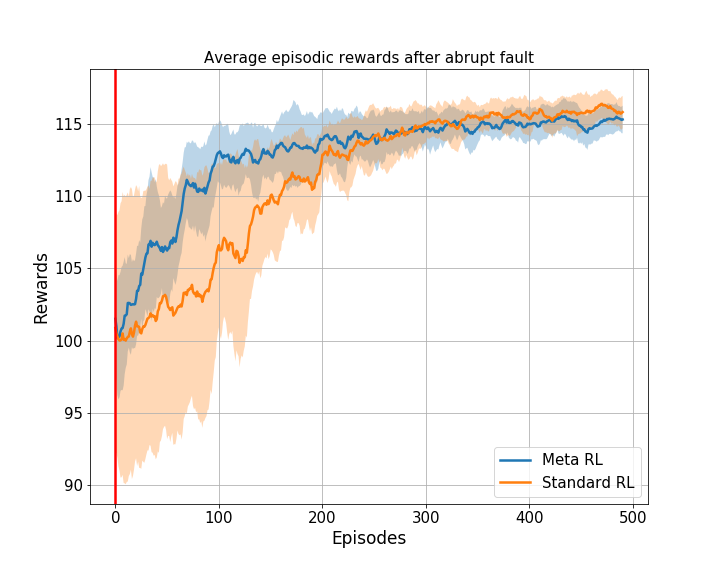} 
\caption{Episodic rewards after abrupt fault when there is a complement of 3 policies available to the controller.}
\label{fig:fullcomplement}
\end{center}
\end{figure}

Figure \ref{fig:nocomplement} shows performance with $\mathcal{C}=\{\varnothing\}$. Episodic rewards start off lower than but comparable to the baseline. They quickly recover and match baseline throughout. Of note is the low variance in episode rewards compared to the baseline. Figure \ref{fig:fullcomplement} shows performance with a complement of 3 policies. The controller starts off with performance similar to the baseline, but quickly pulls ahead and converges to an optimum. The initialization using a populated complement allows the controller to converge to a solution faster.

Additional experiments with different values of learning rates and loop iterations are documented in section \ref{sec:addfigures}.

\section{Related Work} \label{sec:related}


Reinforcement learning has been explored for control systems. \cite{lewis2012reinforcement} surveys RL approaches for feedback control. \cite{liu2016} attempts to speed up learning of neural network policies for controlling systems by manipulating the parameter update rule.


Approaches besides RL are prevelant in the field of FTC. \cite{jiang2006accepting, zhang2003fault} use performance degraded reference models to generate a library of the system under various conditions. Control is transferred to the policy learned for the most similar model in the library.


Meta RL for FTC is a nascent field. Recently, \cite{nagabandi2018learning} used used model-based RL for quickly adapting control to changed system dynamics. They used MAML and a recurrent network as two approaches to develop a meta-update rule for the environment model parameters. In our case, however, we apply MAML towards updating the policy parameters. Alternatively, \cite{saemundsson2018meta} train a model to predict a latent representation of the environment. The latent variable is fed to the agent as a conditioning variable to represent changed dynamics. \cite{wang2016learning} use a recurrent neural network to train a controller on a population of related environments. The controller, being recurrent, has memory of this experience, and therefore learns an internal function to transition between environments as they change.
\section{Conclusion} \label{sec:conclusion}

We have proposed a meta-RL algorithm, which exploits a controller's past experience under faults to initialize parameters for a new policy under a novel abrupt fault. The meta-update can optionally use a data-driven model to mitigate sample inefficiency, or it can fall back to using importance sampling on buffered experiences to evaluate the complement under current conditions. The newly derived parameters are added to the complement if they are divergent enough from the members of the set, thus ensuring a diverse library of behaviors for faster adaption to new faults.

MAML can be sensitive to choice of model architecture, task, and hyperparameters \cite{howtotrainmaml}. This merits further investigation on guarantees of convergence and optimality under faults. MAML can be further incorporated in our approach by using meta-learning to update the data-driven model itself. This should further reduce time taken to learn an updated model and the dependence on the size of the buffered data.

\bibliographystyle{apacite}
\bibliography{references.bib}

\section*{Appendix}

The code and experimental setup for this work can be found at \url{https://git.isis.vanderbilt.edu/ahmedi/airplanefaulttolerance/-/tree/phm2020}.

\subsection{Hyperparameters}

Unless otherwise specified, the following parameters in table \ref{tbl:hyperparameters} were used.

\begin{table}[h]
\caption{Meta-update parameters}
\label{tbl:hyperparameters}
\begin{tabular}{|l|l|}
\hline
Parameter      & Value  \\ \hline \hline
$|\mathcal{M}|$     & 2000   \\ \hline
$\alpha_{in}$  & 0.001  \\ \hline
$\alpha_{out}$ & 0.001  \\ \hline
$K_{in}$ & 2  \\ \hline
$K_{out}$ & 4  \\ \hline
$s$            & 3      \\ \hline
\end{tabular}
\end{table}

Table \ref{tbl:ppoparams} documents parameters used by our implementation of PPO algorithm.

\begin{table}[h]
\caption{Parameters used by the PPO algorithm.}
\label{tbl:ppoparams}
\begin{tabular}{|l|l|}
\hline
Parameter      & Value                                    \\ \hline \hline
Optimizer      & Adam                                     \\ \hline
$\alpha$       & 0.02                                     \\ \hline
$\beta$        & (0.9, 0.999)                             \\ \hline
Epochs         & 5                                        \\ \hline
$t_{update}$     & 2000                                   \\ \hline
Value network  & (64, tanh, 64, tanh, linear, 1)          \\ \hline
Action network & (64, tanh, 64, tanh, 6, linear, sigmoid) \\ \hline
$\gamma$       & 0.99                                     \\ \hline
$\epsilon$     & 0.2                                      \\ \hline
\end{tabular}
\end{table}

\subsection{Additional Figures}\label{sec:addfigures}

\begin{figure}[h]
\begin{subfigure}{.5\textwidth}
  \centering
  \includegraphics[width=.6\linewidth]{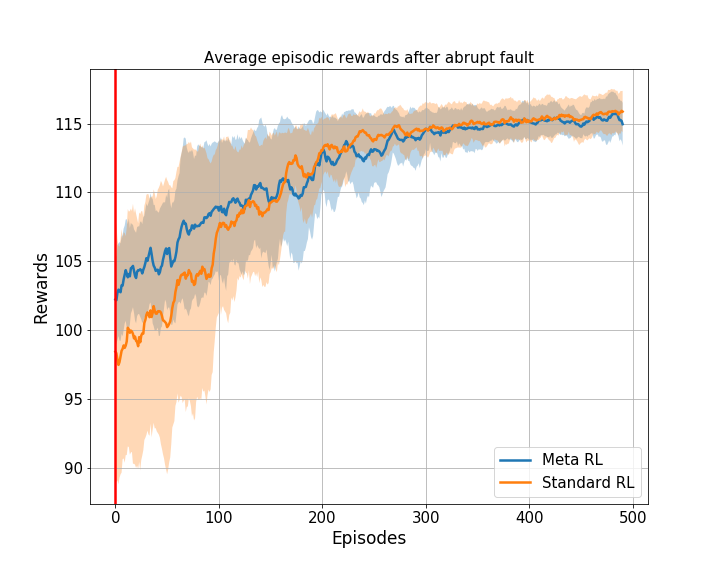}
   \caption{$|\mathcal{C}|=0, \alpha_{in}=0.001, \alpha_{out}=0.001, K_{in}=4, K_{out}=1, p_m=\varnothing$. Fault in tank 4, engine 2.}
\end{subfigure}
\begin{subfigure}{.5\textwidth}
  \centering
  \includegraphics[width=.6\linewidth]{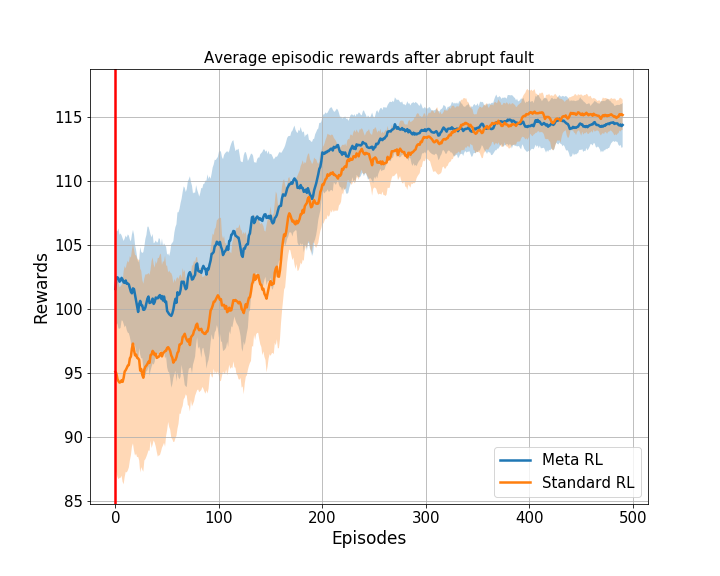}
   \caption{$|\mathcal{C}|=3, \alpha_{in}=0.001, \alpha_{out}=0.001, K_{in}=4, K_{out}=1, p_m=\varnothing$. Fault in tank 4, engine 2.}
\end{subfigure}
\caption{Even with a single meta-update step, $K_{out}=1$, there is noticeable increase in performance.}
\end{figure}

\begin{figure}[h]
\begin{subfigure}{.5\textwidth}
  \centering
  \includegraphics[width=.6\linewidth]{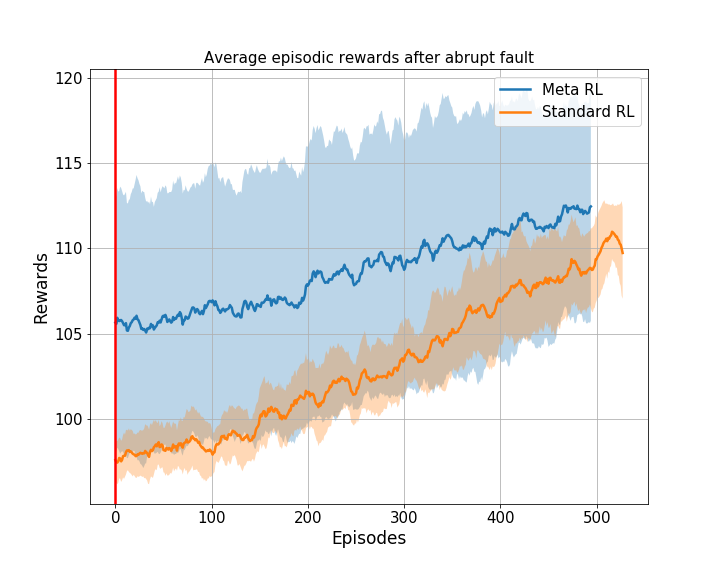}
\caption{$|\mathcal{C}|=0, \alpha_{in}=0.001, \alpha_{out}=0.01, K_{in}=4, K_{out}=2, p_m=\varnothing$. Fault in tank 4, engine 2.}
\end{subfigure}
\begin{subfigure}{.5\textwidth}
  \centering
  \includegraphics[width=.6\linewidth]{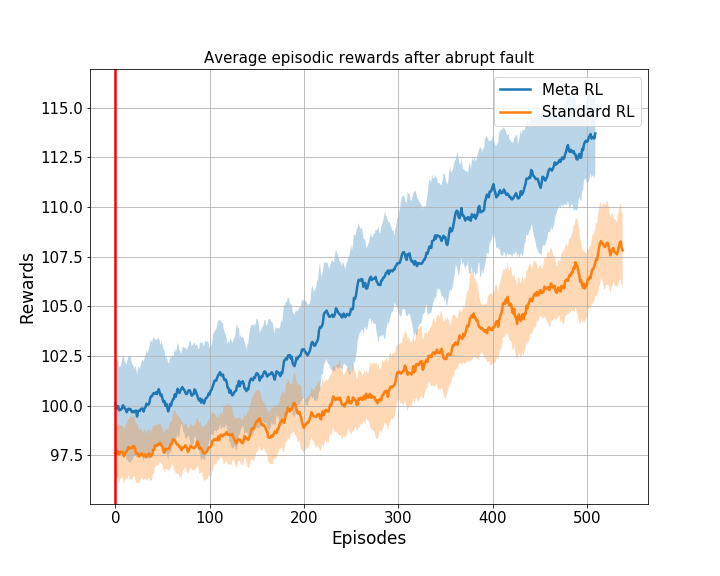}
\caption{$|\mathcal{C}|=3, \alpha_{in}=0.001, \alpha_{out}=0.01, K_{in}=4, K_{out}=2, p_m=\varnothing$. Fault in tank 4, engine 2.}
\end{subfigure}
\caption{Due to a higher outer learning rate $\alpha_{out}$ and iteration number $K_{out}$, the meta-update shows a larger change in performance. With the help of a full complement, the parameter updated is moderated in a direction such that performance variance remains low and shows a higher rate of change.}
\end{figure}

\begin{figure}[h]
\begin{center}
\includegraphics[width=.7\linewidth]{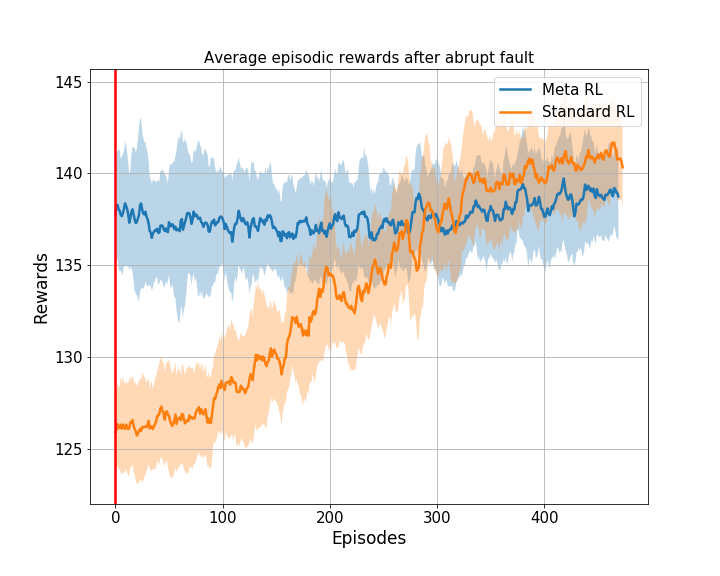}
\caption{$|\mathcal{C}|=3, \alpha_{in}=0.001, \alpha_{out}=0.001, K_{in}=4, K_{out}=2, p_m=\varnothing$. Fault in tank 6. In some faults, the initialization from the meta-update starts at a local optimum.}
\label{}
\end{center}
\end{figure}

\end{document}